\documentclass[12pt]{article}
\usepackage{amsmath}
\usepackage{amsthm}
\usepackage{amssymb}
\usepackage{bm}
\usepackage[round]{natbib}
\usepackage{multirow}
\usepackage{url}
\usepackage[margin=1in]{geometry}

\usepackage[utf8]{inputenc} 
\usepackage[T1]{fontenc}    
\usepackage[hidelinks]{hyperref}       
\usepackage{booktabs}       
\usepackage{amsfonts}       
\usepackage{nicefrac}       
\usepackage{microtype}      
\usepackage{algorithm}
\usepackage[noend]{algpseudocode}
\usepackage{nicefrac}
\usepackage{tikz}
\usepackage{graphicx}
\usepackage{enumitem}
\usepackage{wrapfig}

\newcommand\independent{\protect\mathpalette{\protect\independenT}{\perp}}
\def\independenT#1#2{\mathrel{\rlap{$#1#2$}\mkern2mu{#1#2}}}
\newcommand{\R}{\mathbb{R}}

\makeatletter
\def\BState{\State\hskip-\ALG@thistlm}
\makeatother

\title{Learning non-parametric Markov networks with mutual information}
\date{}
\begin{document}

\author{Janne Leppä-aho$^{1,*}$, Santeri Räisänen$^1$, Xiao Yang, Teemu Roos$^1$ \\
\begin{small}$^{1}$HIIT / Department of Computer Science, University of Helsinki\end{small} \\
\begin{small}$^*$\textit{janne.leppa-aho@helsinki.fi}\end{small}}
\maketitle

\begin{abstract}
We propose a method for learning Markov network structures for continuous data without invoking any assumptions about the distribution of the variables. The method makes use of previous work on a non-parametric estimator for mutual information which is used to create a non-parametric test for multivariate conditional independence. This independence test is then combined with an efficient constraint-based algorithm for learning the graph structure. The performance of the method is evaluated on several synthetic data sets and it is shown to learn considerably more accurate structures than competing methods when the dependencies between the variables involve non-linearities.
\end{abstract}

\section{Introduction}

This paper addresses the problem of learning a Markov network structure from continuous data without assuming any particular parametric distribution. The large majority of the existing methods approach this problem by assuming that the variables follow a multivariate normal distribution. This essentially reduces the problem of learning whether two variables are independent to deciding if they have non-zero partial correlation. However,  as the correlation measures only the strength of a linear dependence, the methods utilizing this might not be able to capture the dependence structure correctly when the relationships are non-linear or the data deviates from the multivariate Gaussian.

To remedy this, we opt to use conditional mutual information to measure the strength of association between the random variables. Like correlation, the mutual information equals zero for independent random variables which makes it possible to use it with Markov network structure learning algorithms based on independence testing, but unlike correlation, mutual information captures any kind of dependence and equals zero only if the variables are independent. In order to compute the mutual information without assumptions about the distributions of variables, we use the non-parametric estimators from previous work \citep{kozachenko1987,kraskov2004,vermeljka2008} which are based on $k$-nearest neighbour statistics.

The literature on methods for non-parametric learning of Markov network structures in the continuous setting is scarce. \cite{Hoffmann97} present an approach which involves approximating the conditional densities of variables with neural networks. One popular semiparametric approach is to assume that there exists univariate transformations for each variable after which the joint distribution of the transformed variables is multivariate normal. This then allows one to use all the machinery developed for Gaussian data. The resulting model class and the methods are termed \textit{non-paranormal} or \textit{Gaussian copulas} \citep{Liu2009,liu2012}. Another related approach is by \cite{Yang2012,Yang2015} where the authors assume that the node-wise conditional distributions belong to exponential family. The graphical model can be then learned by fitting penalized generalized linear models for each of the variables to find their neighbouring nodes. In addition, the distribution free learning of \textit{Bayesian networks} has been studied by \cite{Margaritis2005} and \cite{Sun2008}. 

In Section \ref{sec:MI}, we will review how mutual information is estimated from continuous data based on $k$-nearest neighbour statistics, and how this estimator can be used for testing conditional independence. Section \ref{sec:UG} goes through the constraint-based algorithm which we will use to learn the Markov network structures. In Section \ref{sec:EXP}, we study the performance of our method with several synthetic data sets to illustrate the distinct behaviour of the proposed method especially when the data involves non-linearities. 
       
\section{Independence testing using mutual information}\label{sec:MI}
In this section we present the Kraskov estimator for mutual information and show how it can be used for independence testing. 
\subsection{Preliminaries}
Let $X$ and $Y$ denote two continuous random variables with densities $f_X(x)$ and $f_Y(y)$, respectively. Mutual information \citep{Cover2006} measures the information that one random variable carries about the other and is defined as
\begin{equation}\label{MI}
I(X;Y) = \iint  f_{XY}(x,y)\log\frac{f_{XY}(x,y)}{f_X(x)f_Y(y)}dxdy, 
\end{equation}where $f_{XY}(x,y)$ is the joint density of $X$ and $Y$. We can express mutual information using entropies as
\begin{equation}\label{MIthroughENT}
I(X;Y) = H(X) + H(Y) - H(X,Y),
\end{equation}where $H(\cdot)$ denotes entropy. Let $Z$ be a random vector. Conditional mutual information between $X$ and $Y$ given $Z$ is defined as the expected value 
$$
I(X;Y\mid Z) = \mathbb{E}_{f_{XYZ}} \log \frac{f_{XY\mid Z}(X,Y\mid Z)}{f_{X\mid Z}(X\mid Z )f_{Y\mid Z}(Y\mid Z)}.$$ In terms of entropy, the conditional mutual information decomposes to
\begin{align}
I(X;Y\mid Z) &= H(X,Z) + H(Y,Z) \\ \notag
& - H(X,Y,Z) - H(Z).
\end{align}It is easy to see from (\ref{MI}) that mutual information equals zero if the variables $X$ and $Y$ are independent. The same holds for the conditional mutual information: $$I(X;Y \mid Z) = 0 \Leftrightarrow X \independent Y \mid Z.$$ 

\subsection{Estimating mutual information}
Here, we review how the quantities $H(X), I(X,Y)$ and $I(X;Y\mid Z)$ can be estimated given the observed samples $x_i$, $y_i$ and $z_i$, where $i = 1,\ldots, n$. 

The Kraskov estimator for mutual information builds on the previous entropy estimator by \cite{kozachenko1987}. The derivation of this entropy estimator is presented in \citep{kraskov2004} and it starts from the definition, which can be interpreted as the expected value of $- \log f_X(X)$. This implies that if one has an unbiased estimator for $\log f_X(X)$, then the unbiased estimate for entropy can be obtained as a sample average over local log-probability density estimates. Assuming that the probability density is constant in hyperspheres containing the $k$-nearest neighbours of each data point, one arrives in the following formula:
\begin{equation}\label{KLentropy}
\hat{H}(X) = \psi(n) - \psi(k) + \log c_d + \frac{d}{n}\sum_{i=1}^n \epsilon (i),
\end{equation} where $\epsilon(i)$ is twice the distance to the $k$:th nearest neighbour of data point $x_i$, $\psi(\cdot)$ is the digamma function, $d$ denotes the dimension of $X$ and  $c_d$ is the volume of the unit ball w.r.t. the used norm. From now on, we assume that the maximum norm is used, implying $\log c_d = 0$.

\citeauthor{kraskov2004} expand this to mutual information estimation with help of the formula (\ref{MIthroughENT}). Naively applying the estimate (\ref{KLentropy}) for each of the entropies in (\ref{MIthroughENT}) would induce errors due to the different length scales in spaces $(X,Y)$, $X$ and $Y$. Instead, \citeauthor{kraskov2004} fix the length scale by searching the $k$-nearest neighbours first in the joint space $(X,Y)$. Let $\epsilon(i)/2$ denote the distance to the $k$:th nearest neighbour of the point $(x_i,y_i)$. When computing the entropy estimate in the marginal space $X$, the following approximation is used:
$$\psi(k) = \frac{1}{n}\sum_{i=1}^n \psi(n_x(i) + 1),$$
where $n_x(i)$ is the number of points $x_j$ such that $||x_i - x_j|| < \epsilon(i)/2, \ j\not= i$. The similar approximation is used in the $Y$ space by replacing the $x_i$ with $y_i$. This is motivated by the fact that Eq. (\ref{KLentropy}) holds for any $k$, and $\epsilon(i)/2$ is the distance either to the $(n_x(i) +1)$:th neighbour of $x_i$ or to the $(n_y(i) + 1)$:th neighbour of $y_i$. 
Using equations (\ref{MIthroughENT}) and (\ref{KLentropy}) with the approximation in the marginal spaces leads to the cancellation of the $\epsilon(i)$ terms and we obtain the following formula for the mutual information:     
\begin{equation}\label{KraskovMI}
\hat{I}(X;Y) = \psi(k) + \psi(n) - \frac{1}{n}\sum_{i=1}^n \left( \psi(n_x(i) + 1) + \psi(n_y(i) + 1) \right). 
\end{equation}
Using similar reasoning, \cite{vermeljka2008} present the following formula for the conditional mutual information:
\begin{align}\label{PaulusCMI}
\hat{I}(X;Y \mid Z)& = \psi(k) - \frac{1}{n}\sum_{i=1}^n ( \psi(n_{xz}(i)+1)  + \psi(n_{yz}(i)+1) - \psi(n_z(i) + 1) ),
\end{align}where the counts $n_z(i), n_{yz}(i)$ and $n_{xz}(i)$ in the marginal spaces are defined in a similar fashion as in Eq. (\ref{KraskovMI}).

The parameter $k$ in these estimators controls the bias-variance trade-off: a small $k$ means that the assumption about the constant density holds only in small regions, thus implying smaller bias, whereas large $k$ decreases the variance as more data are used to obtain the local estimates. In our experiments, we set $k = 3$ as suggested in \citep{kraskov2004}. Even though this suggestion concerns the estimator (\ref{KraskovMI}) and it is not evident how this translates to the conditional estimator, this seemed to produce satisfactory performance in all our experiments. 

\subsection{Non-parametric test for conditional independence}
Due to statistical variation, the empirical joint distribution is
hardly ever exactly equivalent to the product of the margins,
just like an empirical correlation coefficient is hardly ever
exactly zero. Hence, we need to consider a test that takes
into account the statistical uncertainty of the mutual
information estimator. To this end, we apply a permutation
test to simulate the sampling distribution of the mutual
information statistic under the null hypothesis of conditional
independence.

To test the conditional independence based on observed data $\bm{x}$, $\bm{y}$ and $\bm{z}$, we first set a significance level $\alpha$, and compute the estimate $\hat{I}(\bm{x};\bm{y}\mid\bm{z})$. Then,
conditional independence is simulated by randomly permuting the samples $\bm{y} = (y_1, \ldots , y_n)$ and computing $\hat{I}(\bm{x};\bm{y}_{perm(i)}\mid\bm{z})$. This is repeated $T$ times. After this, we count the number of permuted mutual information values that are greater than or equal to the initial estimate $\hat{I}(\bm{x};\bm{y}\mid\bm{z})$. We let $K$ to denote this number. This gives us an estimate for the $p$-value, $\hat{p} = (K+1)/(T + 1)$, which is then compared to the significance level $\alpha$. To ease the computational burden, we also defined a threshold so that if the value for the estimated conditional mutual information fell below $0.001$ nats and the partial correlation based test accepted independence (with the same significance level $\alpha$), then the permutation tests were skipped. 

A similar kind of permutation test was used with mutual information in the context of feature selection in \cite{francois06}. Pseudocode for the conditional independence test is presented in Algorithm \ref{alg:permutationTest}.  

\begin{algorithm}[tb]
\caption{Conditional Independence Test}\label{alg:permutationTest}
\begin{algorithmic}[1]
\Require
      \Statex Significance level $\alpha$, number of iterations $T$
\Procedure{CIT}{$\bm{x},\bm{y},\bm{z}$}
\State $estCMI \gets \hat{I}(\bm{x};\bm{y}\mid\bm{z})$
\State $\textit{PermutedMI} \gets \emptyset $
\For{$i\gets 1, T$}
\State $\bm{y}_{perm(i)} \gets \text{random permutation of }\bm{y}$
\State $mi \gets \hat{I}(\bm{x};\bm{y}_{perm(i)}\mid\bm{z})$
\State $\textit{PermutedMI} \gets \textit{PermutedMI} \cup \{mi\}$
\EndFor
\State $K \gets \#\{\hat{I}(\bm{x};\bm{y}_{perm(i)}\mid\bm{z}) \ \vert \ \hat{I}(\bm{x};\bm{y}_{perm(i)}\mid\bm{z}) \geq estCMI  \}$
\If {$(K+1)/(T + 1) < \alpha$} 
\Return False
\EndIf
\State \Return True
\EndProcedure
\end{algorithmic}
\end{algorithm}

\section{Structure learning of Markov networks}\label{sec:UG}
In this section, we will go briefly through the basic concepts related to Markov networks and then present the structure learning algorithm which is combined with the presented non-parametric conditional independence test. For a more thorough treatment, we refer to \citep{WHITTAKER, LAURITZEN, koller2009}.
   
\subsection{Representation}
Let $X = (X_1, \ldots X_p)$ be a random vector and $G = (V,E)$ denote an undirected graph (UG), where $V = \{ 1, \ldots , p \}$ is the set of nodes corresponding to elements of $X$ and $E \subset V \times V$ the set of edges. Given an UG $G$, we define the Markov blanket of the node $i$ (or variable $i$, these terms are used interchangeably in our context) to be the set containing its neighbouring nodes in the graph $G$, $mb(i) = \{ j \in V \vert (i,j) \in E \}$, where $(i,j) = (j,i)$ is an undirected edge between nodes $i$ and $j$. The graph $G$ encodes a set of conditional independence assumptions that can be characterized via \textit{Markov properties}: 1) the variable $X_i$ is independent of $X_j$ given the remaining ones $V \setminus \{i,j \}$ if $(i,j) \not\in E$, 2) every variable $i\in V$ is conditionally independent of all the other variables given its Markov blanket, 3) for the disjoint subsets of variables, $A,B,C \subset V$, it holds that $X_A$ is conditionally independent of $X_B$ given $X_C$ if $C$ separates $A$ and $B$ in the graph. The notation $X_A$ stands for the random vector containing the variables belonging to a set $A\subset V$. These properties are termed the pairwise, the local and the global Markov properties, respectively. 

A strictly positive distribution $p(X)$ which satisfies the Markov properties implied by the graph $G$ (and only those) also factorizes according to the cliques of $G$ as $p(X) = Z^{-1}\prod_{C\in\mathcal{C}}\phi_C(X_C)$, 
where the functions $\phi_C:\R^{|C|}\to\R_{+}$ are called clique potentials and $Z$ is the normalizing constant.  A clique is a completely connected subset of $V$.
\subsection{Structure learning}
The main problem we are focusing on here is learning the graph structure $G$ based on the observed data $\bm{X} = (\bm{x}_1, \ldots , \bm{x}_n)$, where $\bm{x}_i \in \R^p$ is i.i.d sample from the distribution $p(X)$. The methods addressing this problem are usually either score- or constraint-based ones. The first mentioned approach is based on a data-dependent scoring function which evaluates the goodness of different structures whereas the constraint-based methods make use of the Markov properties and perform a series of conditional independence tests to infer the network structure \citep{Schluter2014}. Here, we will adopt this latter approach.   

More in detail, we will use the IAMB algorithm \citep{Tsamardinos03} to learn the Markov blanket for each of the nodes. This algorithm constructs the blanket by first adding variables (with the highest conditional mutual information) until the node under consideration is conditionally independent of all the other given the current blanket. This is followed by a step where variables are removed if they are conditionally independent of the target node given the remaining variables in blanket. The algorithm is guaranteed to return the correct Markov blanket assuming faithfulness and correctness of the independence tests \citep{Tsamardinos03}. The original paper actually assumes that the distribution is faithful to some directed acyclic graph. However, the IAMB is also correct when the underlying graph is undirected as shown in \citep{PENA2007}.     

For any finite sample size $n$, the found Markov blankets are not necessarily coherent in a sense that $i \in mb(j)$ would imply that $j$ was also found to belong to Markov blanket of $i$. To overcome this we define the estimated undirected graph using conservative AND-rule, meaning that there is an undirected edge between $i$ and $j$ if $i \in \widehat{mb}(j)$ and $j \in \widehat{mb}(i)$.    

Implementing this algorithm with the non-parametric independence test described in Section \ref{sec:MI} yields our proposed method, which will be henceforth referred to as \texttt{knnMI\_AND}.

\subsection{On computational complexity}
The computational cost of our proposed approach is dominated by the nearest neighbours searches which become costly especially when the dimension of the data grows. In the concrete implementation of the algorithm we use $kd$-tree \citep{Bentley1975} to perform these queries. Let us analyse the steps needed to compute the estimate for conditional mutual information defined in (\ref{PaulusCMI}). Let $n$ be the number of observations and $d$ denote the dimension of the joint space $(X,Y,Z)$. 
\begin{itemize}
\item[1.] Index construction for joint and marginal spaces takes $O(dn\log n)$ time.
\item[2. ]For each data point $d_i$, we need to find the $k$-nearest neighbour in the joint space and record the distance $\epsilon_i/2$. For a fixed $d$, finding one neighbour has expected running time of $O(\log n)$ \citep{Friedman1977}, which yields a total running time of $O(kn\log n)$. However, with respect to dimension $d$ the time complexity is exponential.
\item[3.] Using the found distances, we count for each data point the number of points whose distance is less than $\epsilon_i/2$. This is done in spaces $(X,Z)$,$(Y,Z)$ and $Z$. With fixed $d$ this would naively take $O(n^2)$ time. 
\end{itemize}

In practice, when the dimension is fixed, the expected running times for single nearest-neighbour and radius queries in $kd$-trees could be significantly smaller, even a constant time operations \citep{Bentley1990}. 

The number of independence tests and association computations (in our case estimating the conditional mutual information) performed by IAMB when searching for a single Markov blanket is in the worst case of order $O(p^2)$ \citep{Tsamardinos03}. However, the authors state they experimentally observed an average case order of $O(p|mb(i)|)$ tests, where $|mb(i)|$ refers to the size of the Markov blanket for some variable $i$. This implies that in the worst case finding the graph takes $O(p^3)$ tests but if the Markov blankets are relatively small, the complexity is considerably lower. 


\section{Experiments}\label{sec:EXP}
In this section we evaluate the performance of the proposed approach and compare it two other methods by creating synthetic data from various Markov network structures where the dependencies between the variables are not necessarily linear or the distribution close to multivariate normal\footnote{The code to reproduce all the experiments is available at \url{https://github.com/janlepppa/graph_learn_mi}}.   

\subsection{Considered methods}
We compare the performance of \texttt{knnMI\_AND} to a method that uses exactly the same structure learning algorithm but with an independence test based on Fisher's z-transformed sample partial correlations, see, for instance, \citep{kalisch2007}. We will refer to this method as \texttt{fisherZ\_AND}. Other methods we compare against include graphical lasso (\texttt{glasso}) \citep{GLASSO1} and neighbourhood-selection method (\texttt{mb}) by \cite{MB06}. All the three previously mentioned methods are based on the multivariate normal assumption. 

The \texttt{glasso} method learns the graph by estimating the inverse of covariance which is done by optimizing an objective function comprising of $\ell_1$-penalized Gaussian log-likelihood. The \texttt{mb} estimates the graph by conducting $\ell_1$-penalized linear regression independently for each variable to find their Markov blankets. We will use the similar AND-rule as mentioned before to construct the graph from the estimated Markov blankets. As the output of \texttt{glasso} and \texttt{mb} depends on the tuning parameter $\lambda > 0$ which controls the amount of $\ell_1$-regularization, we computed graphs for $20$ tuning parameter values, starting from the tuning parameter value $\lambda_{max}$ that resulted in an empty graph and then decreased it to a value $\lambda_{min} = 0.01\lambda_{max}$. The densest model had always more edges than the true generating network structure. The best model was chosen according to the StARS criterion \citep{STARS}. With \texttt{mb}, we tried also choosing the parameter automatically as proposed by the authors to be $\lambda = (n^{-1/2})\Phi^{-1}(1- \alpha/ (2p^2)),$ where $\alpha = 0.05$ and $\Phi(\cdot)$ denotes the c.d.f. of a standard normal random variable. We will refer to this method as \texttt{mb\_auto}. In the experiments, we used the implementations of \texttt{glasso} and \texttt{mb} found in R-package 'huge'\footnote{\url{https://CRAN.R-project.org/package=huge}}. As we mainly study non-Gaussian data, all the input data are put through a non-paranormal transformation based on a shrunken empirical cumulative distribution function (ECDF) \citep{Liu2009,liu2012} before applying \texttt{glasso} or \texttt{mb}. In both, \texttt{knnMI\_AND} and \texttt{fisherZ\_AND}, we set the significance level to be $0.05$. With \texttt{knnMI\_AND} we do $200$ permutations of data when testing for independence.   

To compare the methods, we measure the average Hamming distance (the sum of false positive and false negative edges) between the estimated graph and the ground truth graph.

\subsection{Small network}\label{sec:smallNetwork}
\begin{figure}[tb]
\begin{center}
\includegraphics[width = 0.31\columnwidth]{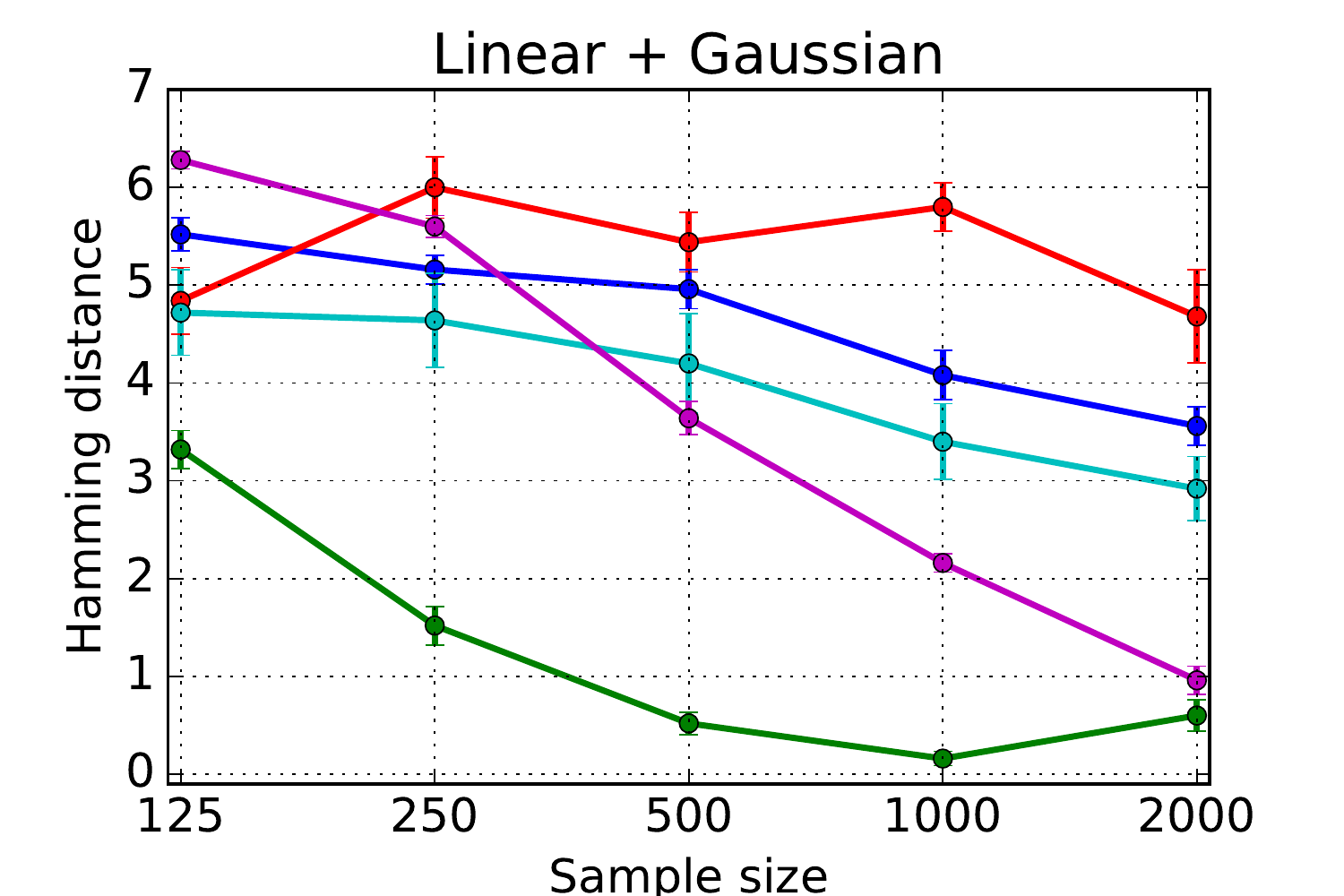}
\includegraphics[width = 0.31\columnwidth]{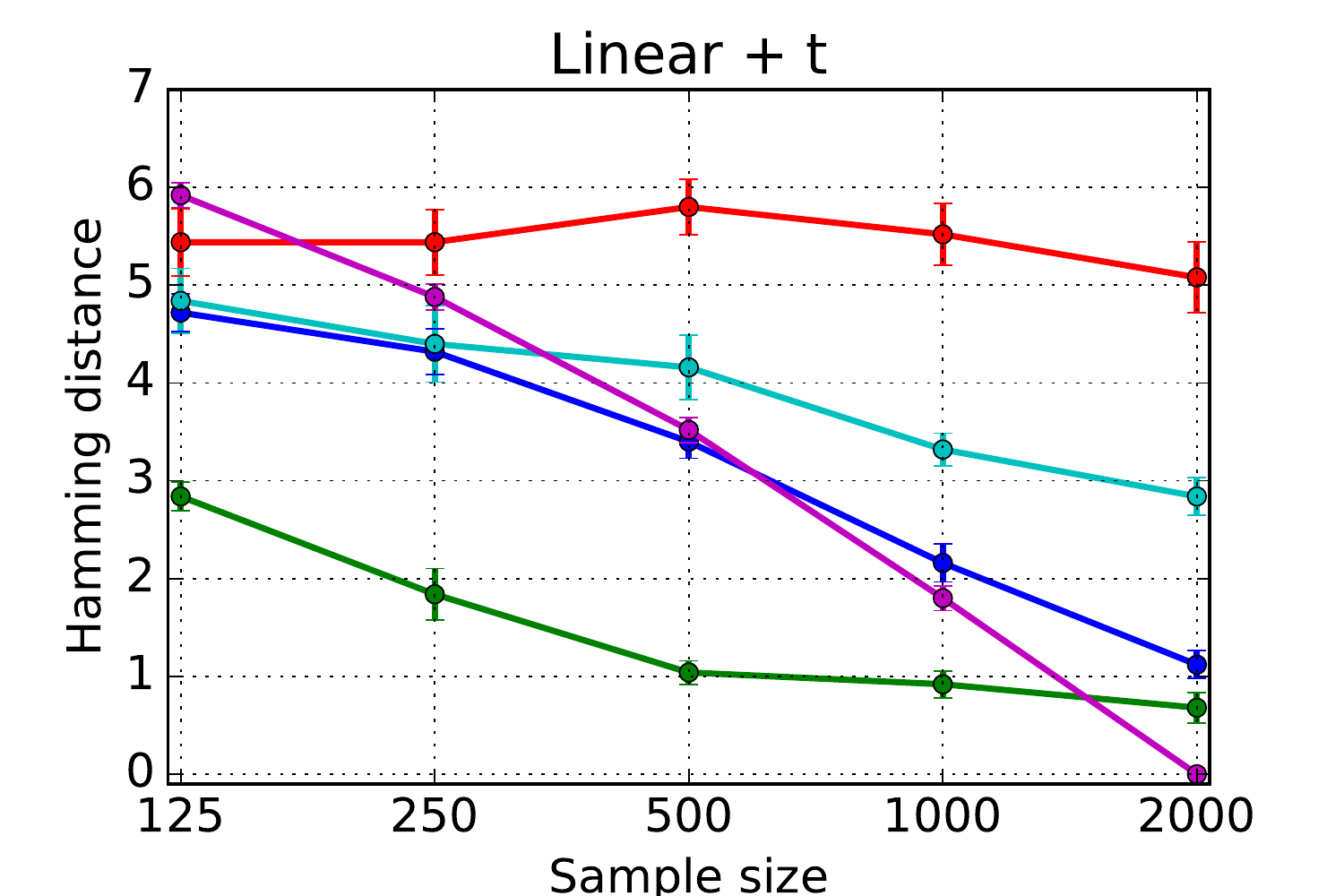}
\includegraphics[width = 0.31\columnwidth]{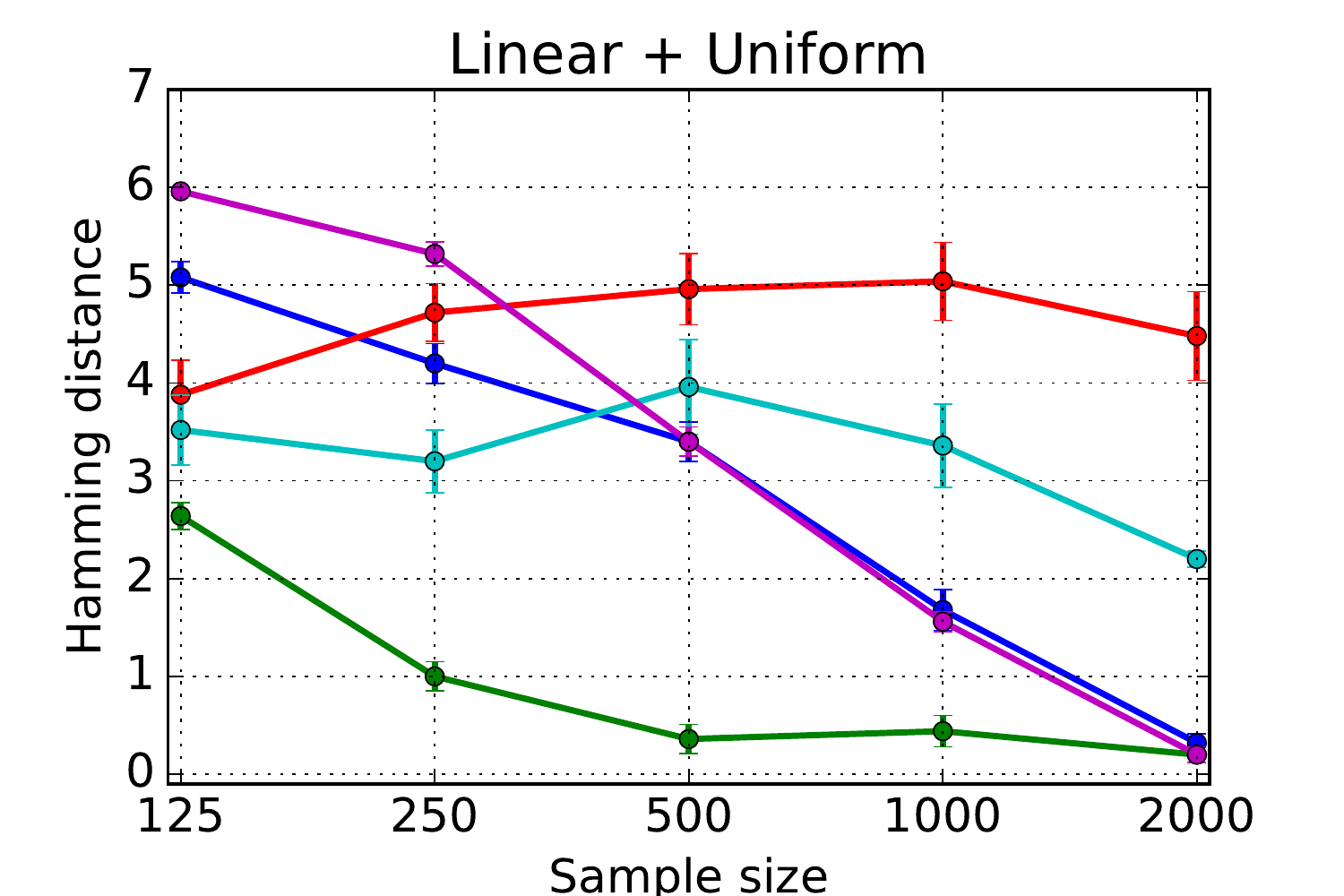} \\
\includegraphics[width = 0.31\columnwidth]{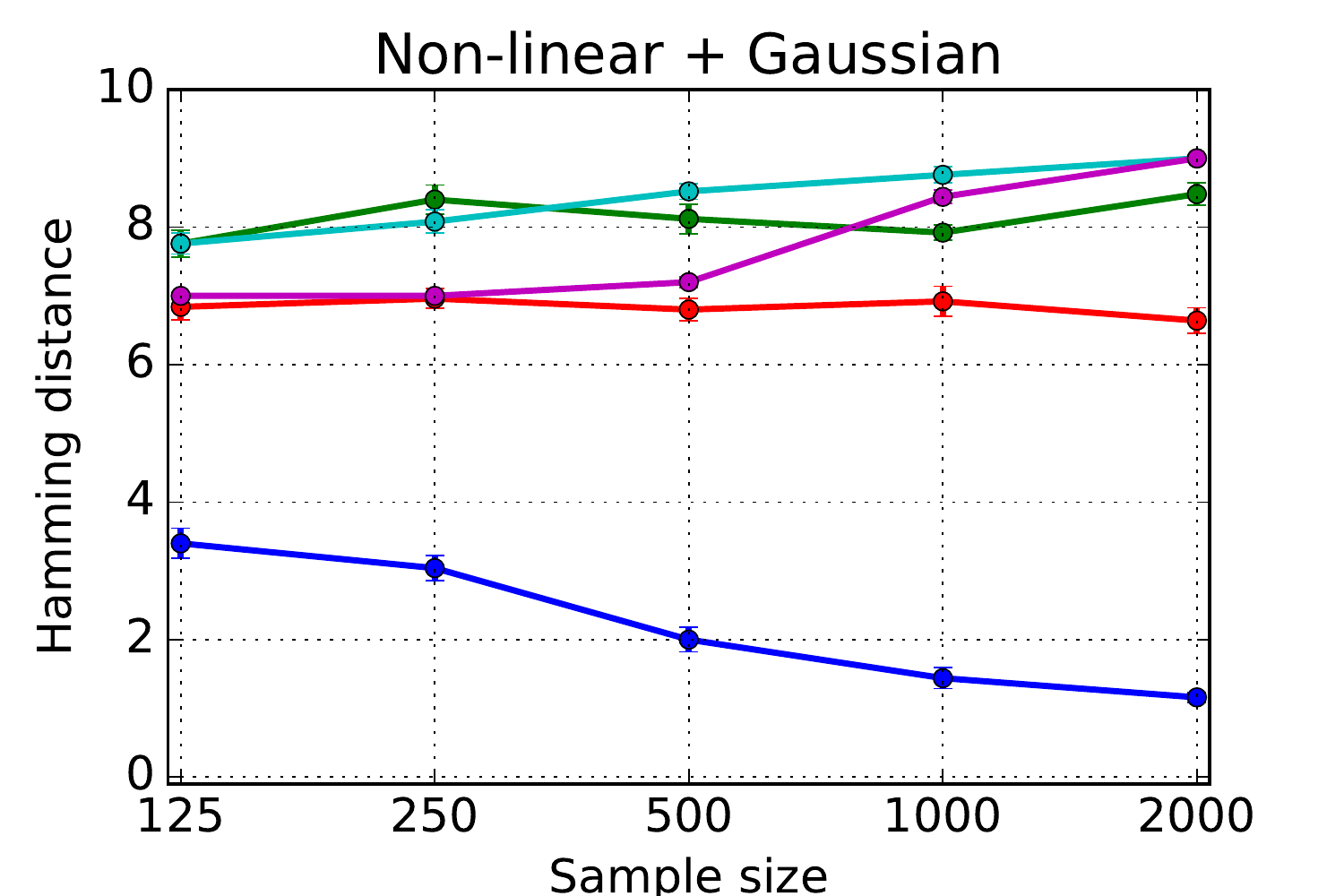}
\includegraphics[width = 0.31\columnwidth]{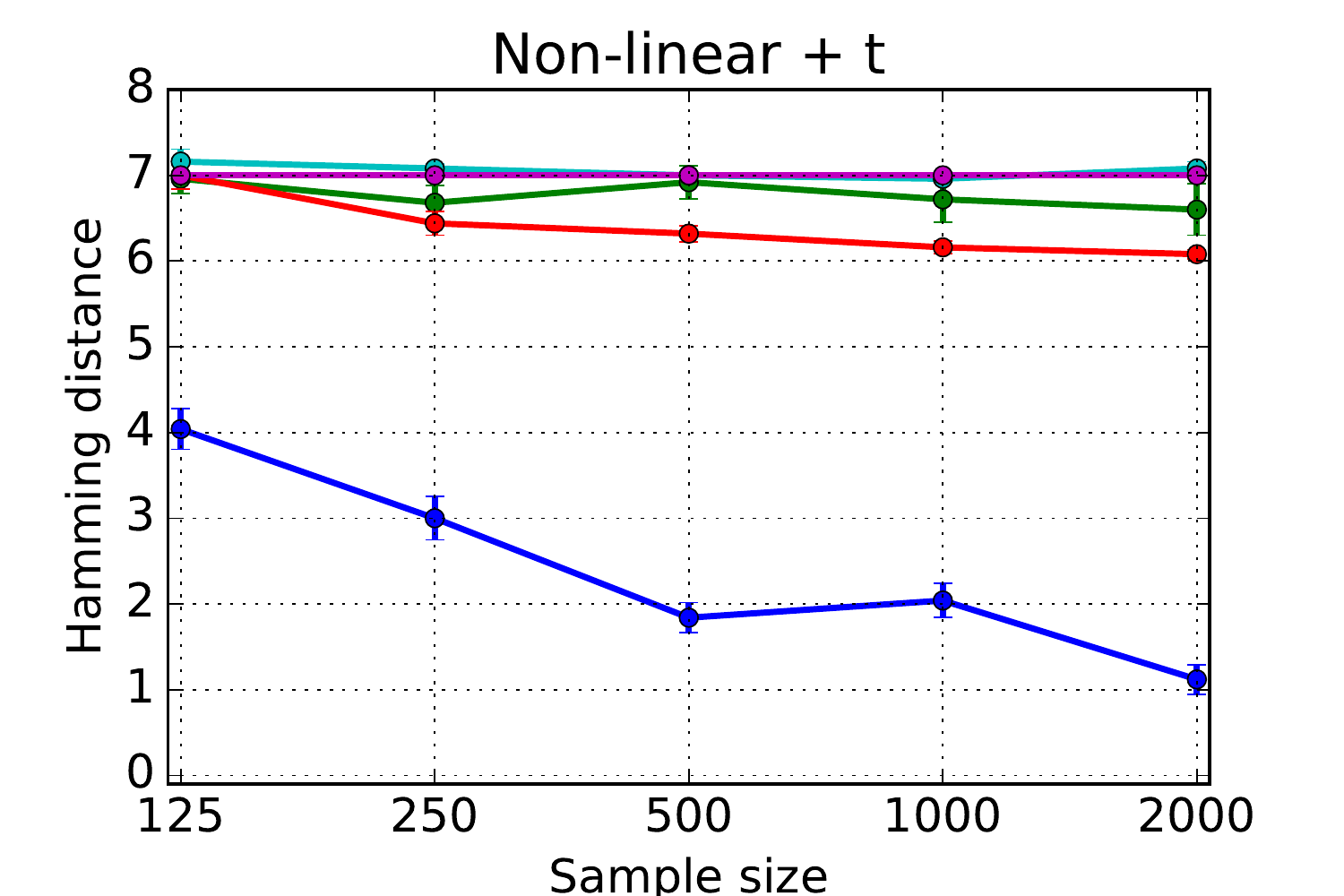}
\includegraphics[width = 0.31\columnwidth]{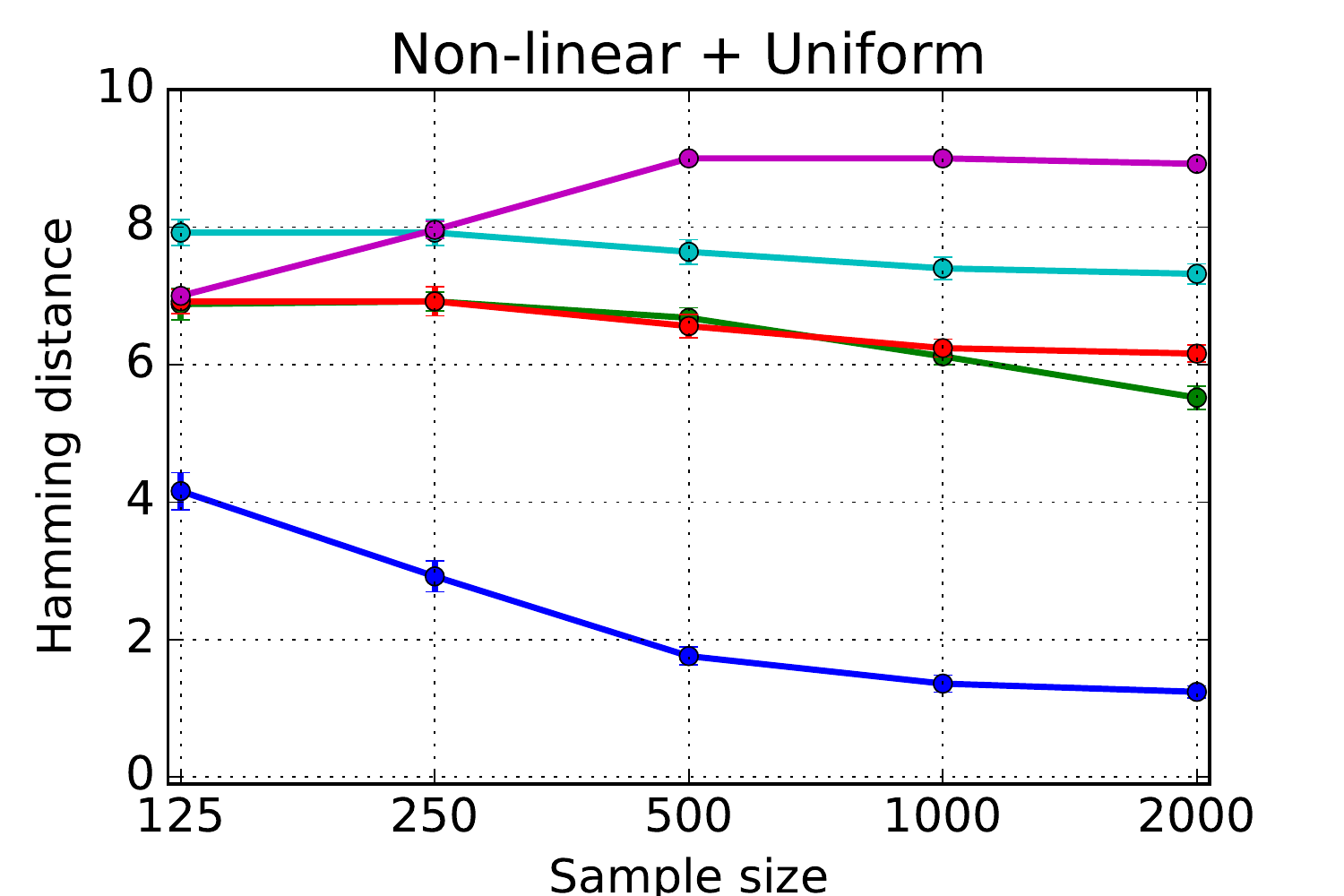}
\includegraphics[width = 0.65\columnwidth]{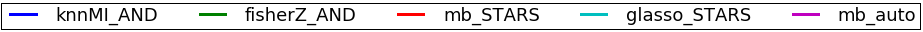}
\end{center}
\caption{Hamming distances for the small network with different noise distributions}\label{smallNetwork}
\end{figure}

First, we consider a small network consisting of seven nodes. In this example, the considered graph is decomposable, implying that we can represent it equally well as a DAG which simplifies the data generation. The ground truth undirected graph and the corresponding DAG are depicted in Figure \ref{fig:smallNetwork}.
With the network structure fixed, we considered six different data generating schemes. The dependencies between the child variable and the parents were either linear or non-linear with an additive noise term. The distribution of noise was selected between standard Gaussian, uniform on the interval $[-1,1]$ and standard $t$ with degrees of freedom set to $2$. The data generating mechanism is presented in Table \ref{tab:smallNetwork}. We use $\epsilon_i$ to denote the noise term which follows one of the aforementioned distributions. 
\begin{wrapfigure}[1]{l}{0.2\textwidth}
\centering
\begin{tikzpicture}[thick, scale=0.55, every node/.style={scale=0.6}]
    \tikzstyle{edge} = [draw,thick,->]
        
        \node[circle,draw] (0) at  (-2.5, 0.5){X1}; 
        \node[circle,draw] (1) at  (-0.5, 0.5){X2}; 
        \node[circle,draw] (2) at  (-1.5, -1){X3}; 
        \node[circle,draw] (3) at  (-1.5, -2.5){X4}; 
        \node[circle,draw] (5) at  (1, 0.5){X6}; 
        \node[circle,draw] (4) at  (1, -1){X5}; 
        \node[circle,draw] (6) at  (1, -2.5){X7};        
		\draw[-,thick] (0) edge (1);
		\draw[-,thick] (1) edge (2);
		\draw[-,thick] (2) edge (3);
		\draw[-,thick] (1) edge (4);
		\draw[-,thick] (2) edge (4);
		\draw[-,thick] (4) edge (5);
		\draw[-,thick] (2) edge (6);
		\draw[-,thick] (4) edge (6);
\end{tikzpicture}
\\\hspace{1cm}
\begin{tikzpicture}[thick, scale=0.55, every node/.style={scale=0.6}]
    \tikzstyle{edge} = [draw,thick,->]
        
        \node[circle,draw] (0) at  (-2.5, 0.5){X1}; 
        \node[circle,draw] (1) at  (-0.5, 0.5){X2}; 
        \node[circle,draw] (2) at  (-1.5, -1){X3}; 
        \node[circle,draw] (3) at  (-1.5, -2.5){X4}; 
        \node[circle,draw] (5) at  (1, 0.5){X6}; 
        \node[circle,draw] (4) at  (1, -1){X5}; 
        \node[circle,draw] (6) at  (1, -2.5){X7};        
		\draw[->,thick] (0) edge (1);
		\draw[->,thick] (1) edge (2);
		\draw[->,thick] (2) edge (3);
		\draw[->,thick] (1) edge (4);
		\draw[->,thick] (2) edge (4);
		\draw[->,thick] (4) edge (5);
		\draw[->,thick] (2) edge (6);
		\draw[->,thick] (4) edge (6);
\end{tikzpicture}
\caption{The small network}\label{fig:smallNetwork}
\end{wrapfigure}
\begin{table}[H]
\caption{Data generating models}\label{tab:smallNetwork}
\begin{flushright}
\begin{tabular}{l  l  l }
\phantom{} &Linear &Non-linear \\
$X_1$ & $\epsilon_1$ 						& $\epsilon_1$ \\
$X_2$ & $0.2X_1 + \epsilon_2$				& $2\cos(X_1)+ \epsilon_2$ \\
$X_3$ & $0.5X_2 + \epsilon_3$		& $2\sin(\pi X_2) + \epsilon_3$	\\
$X_4$ & $0.25X_3 + \epsilon_4$ 			& $3\cos(X_3) + \epsilon_4$ \\
$X_5$ & $0.35X_2 + 0.55X_3 + \epsilon_5$	& $0.75X_2 X_3 + \epsilon_5$ \\
$X_6$ & $0.65X_5 + \epsilon_6$ 						& $2.5X_5 + \epsilon_6$ \\
$X_7$ & $0.9X_3 + 0.25X_5 + \epsilon_7$	& $3\cos(0.2X_3) + \log |X_5| + \epsilon_7$ \\
\end{tabular}
\end{flushright}
\end{table}
\vspace{12pt}
We created multiple data sets with sample sizes ranging from $125$ to $2000$. The average Hamming
distances to the true graph for each method are presented in Figure \ref{smallNetwork}. All the presented values are averages from $25$ repetitions. In the Hamming distance figures, errors bars show the standard error of the mean.     

In the linear case, \texttt{fisherZ\_AND} is the most accurate accurate regardless the noise distribution. It is also somewhat surprising how the performance of \texttt{fisherZ\_AND} did not seem to deteriorate at all when the assumption about normally distributed noise was violated. In the linear case, we can see that \texttt{fisherZ\_AND} and \texttt{mb\_auto} are able to capture the dependencies with smaller amount of samples, especially in the Gaussian case. As maybe expected, our method can learn the structure clearly the best in cases where the dependences are non-linear with a wide margin to other approaches. In these cases, \texttt{knnMI\_AND} is the only method that steadily improves its performance as the sample size increases, recovering the true generating structure almost correctly when sample size $n = 2000$.
\subsection{Non-paranormal data from random networks}
\begin{figure}[tb]
\begin{center}
\includegraphics[width = 0.3\columnwidth]{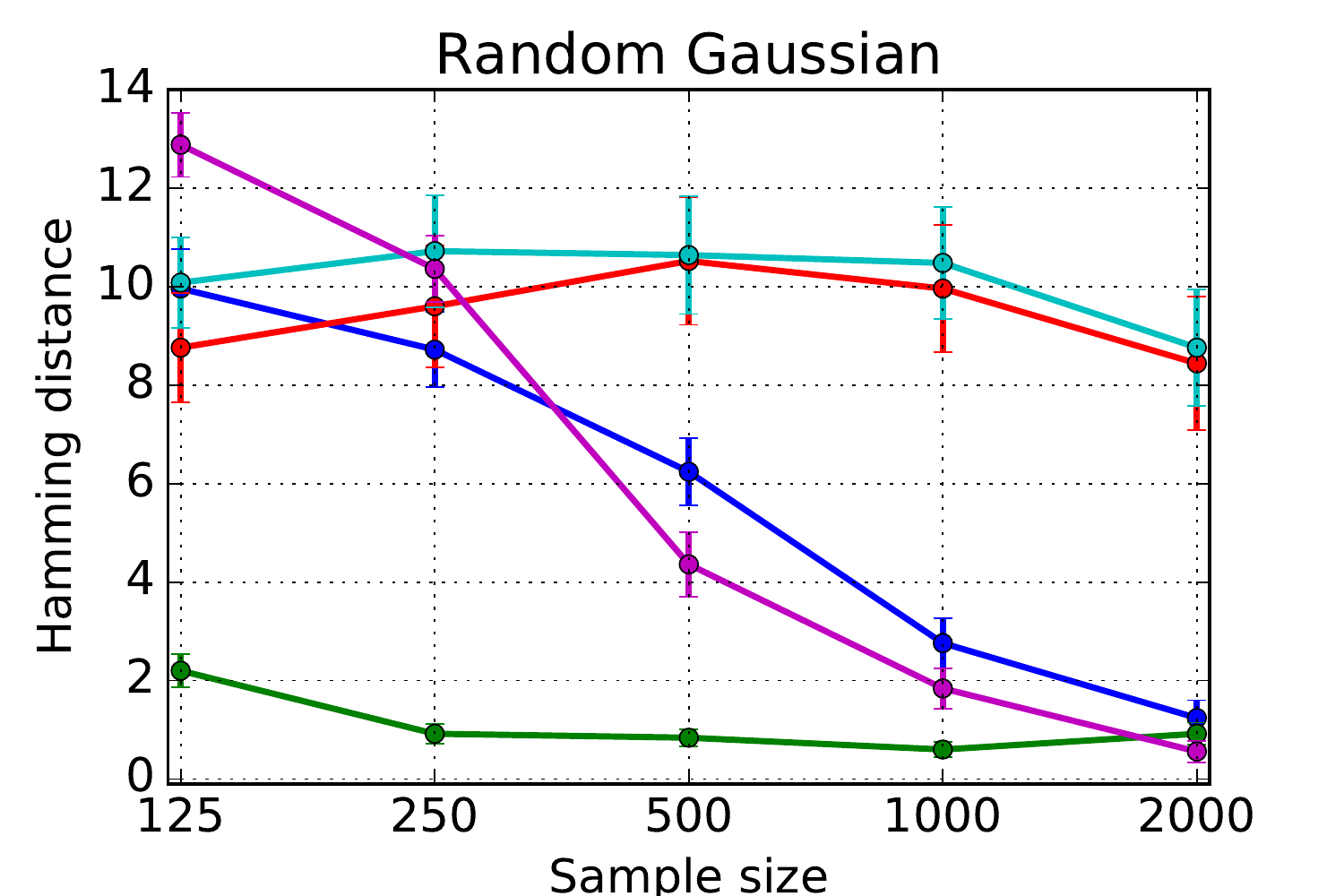}\includegraphics[width = 0.3\columnwidth]{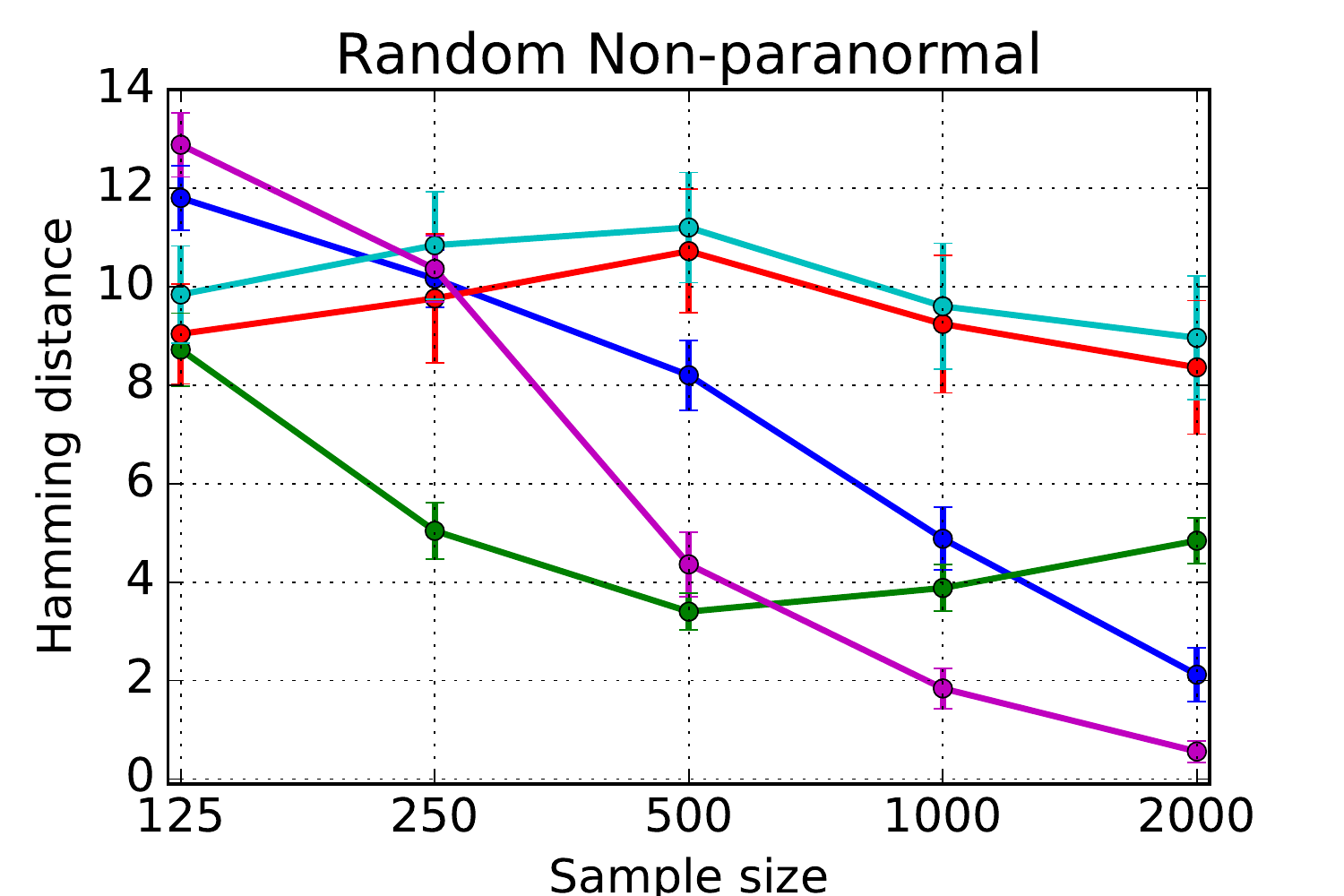}\includegraphics[width = 0.3\columnwidth]{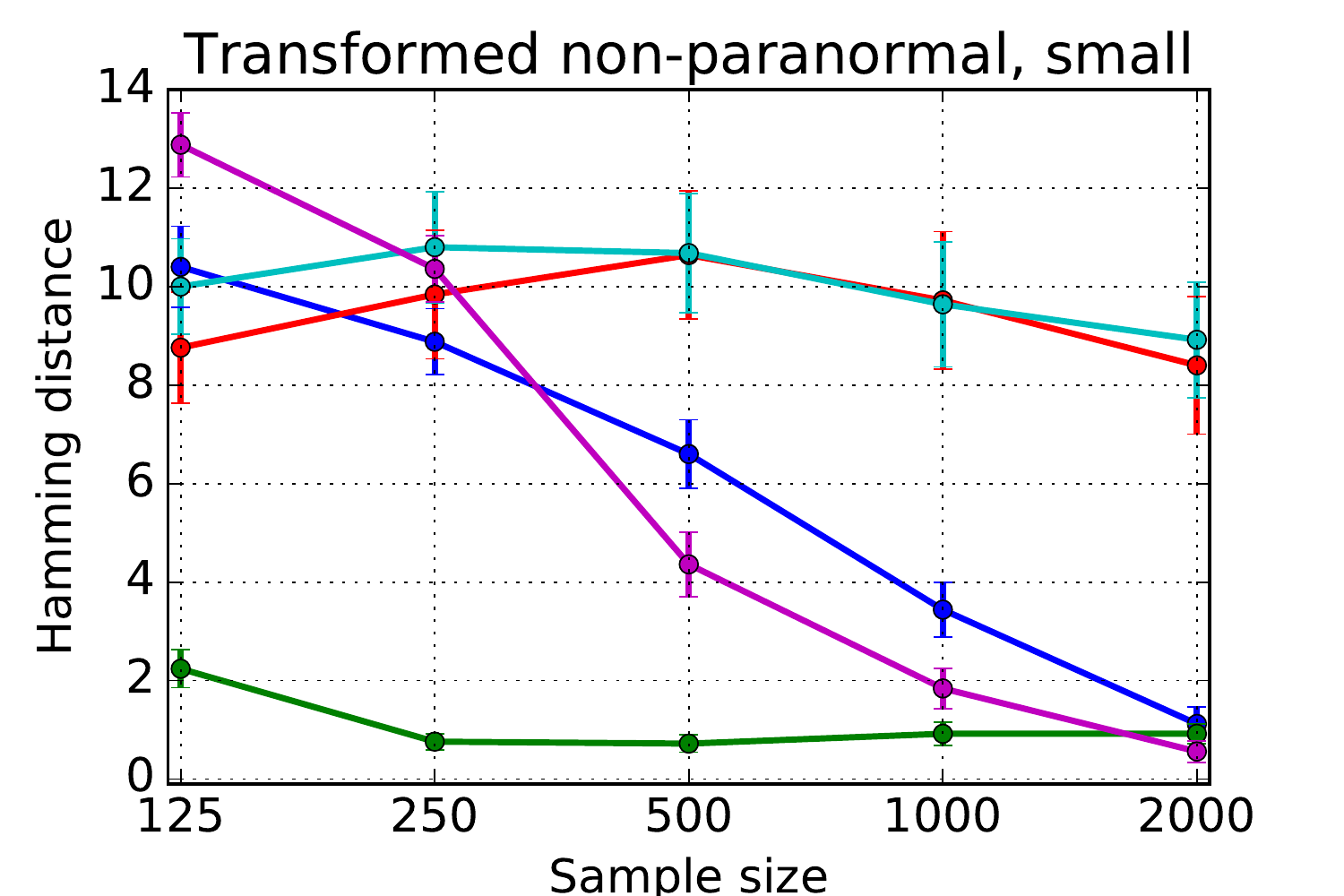} \\
\includegraphics[width = 0.3\columnwidth]{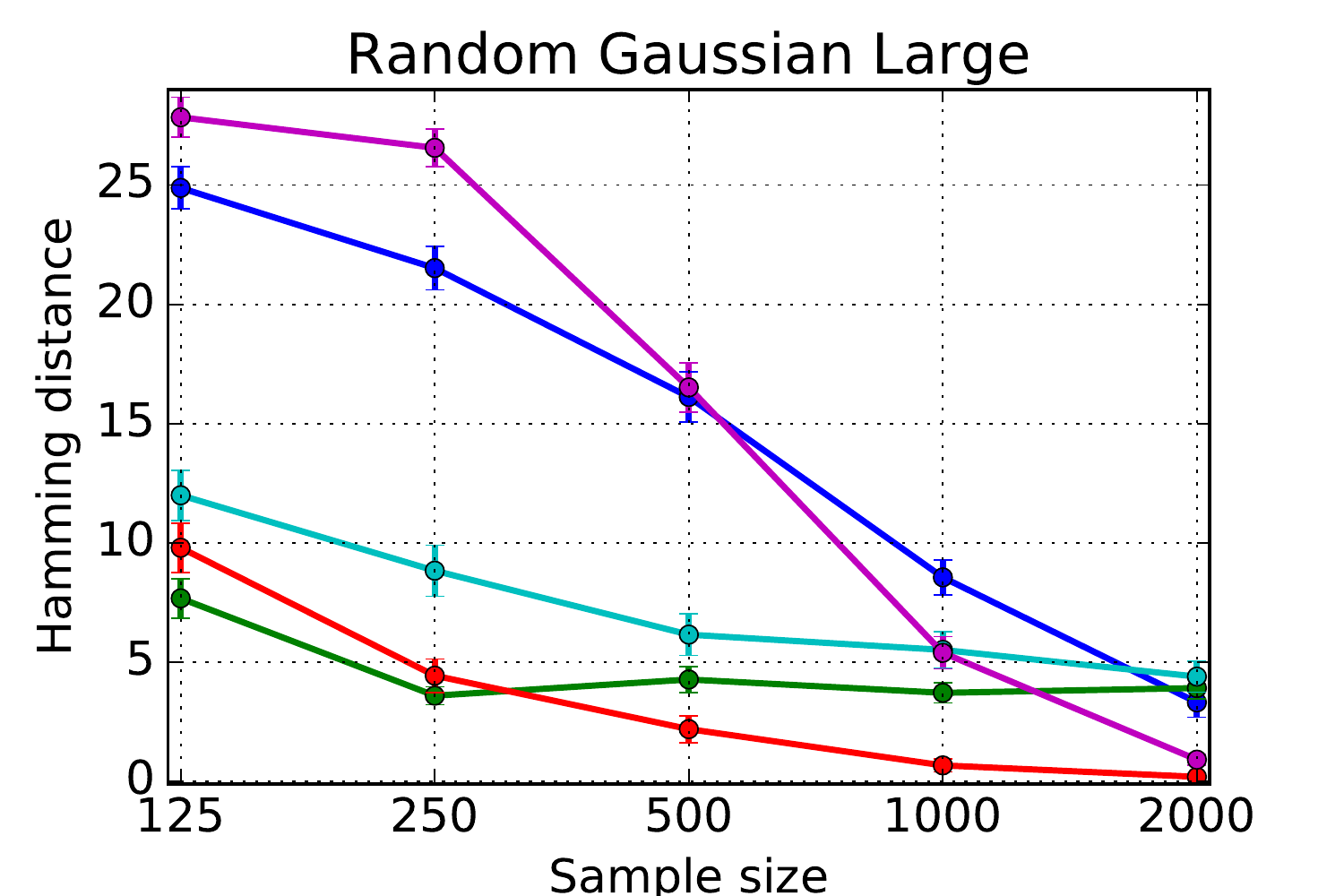}\includegraphics[width = 0.3\columnwidth]{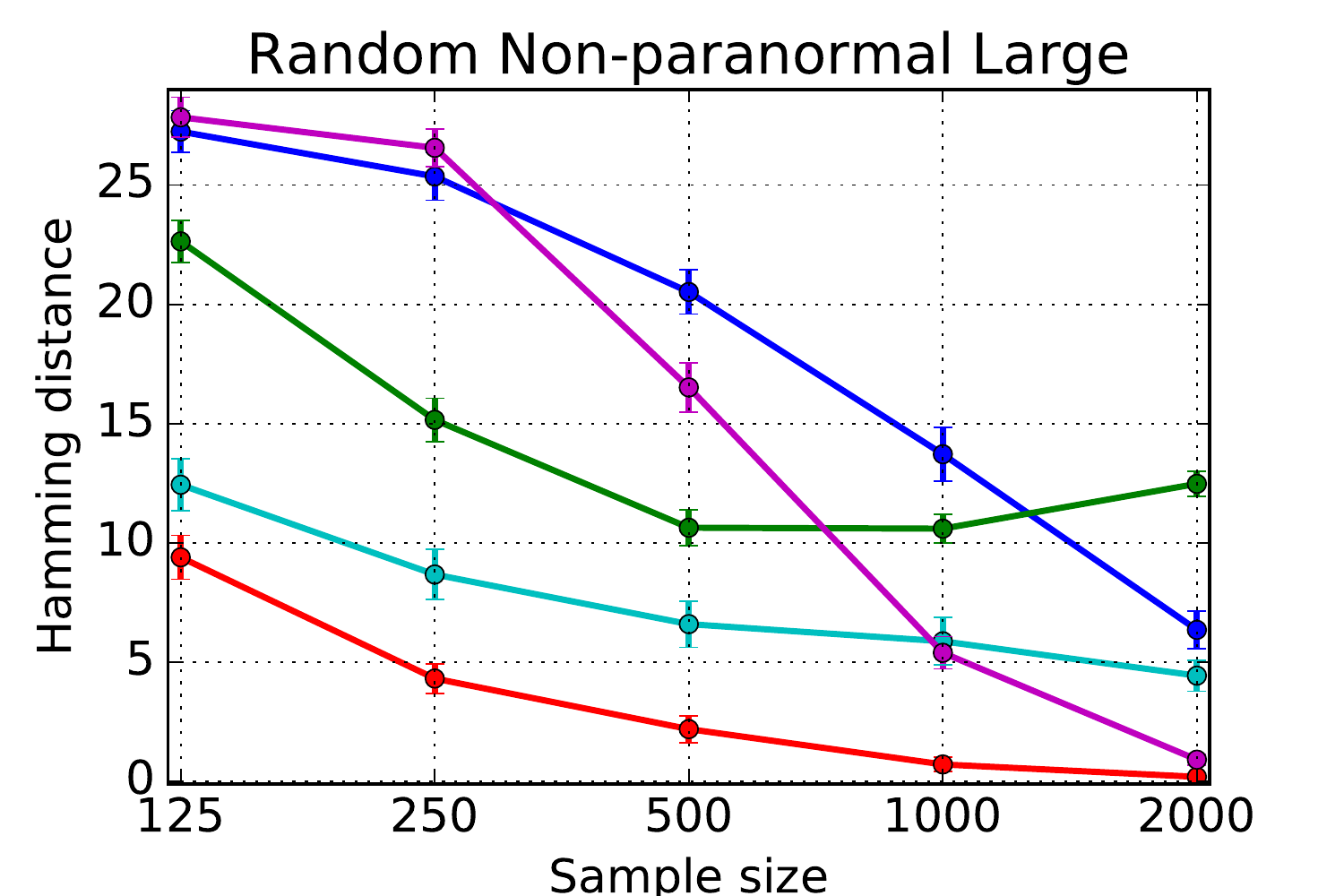}\includegraphics[width = 0.3\columnwidth]{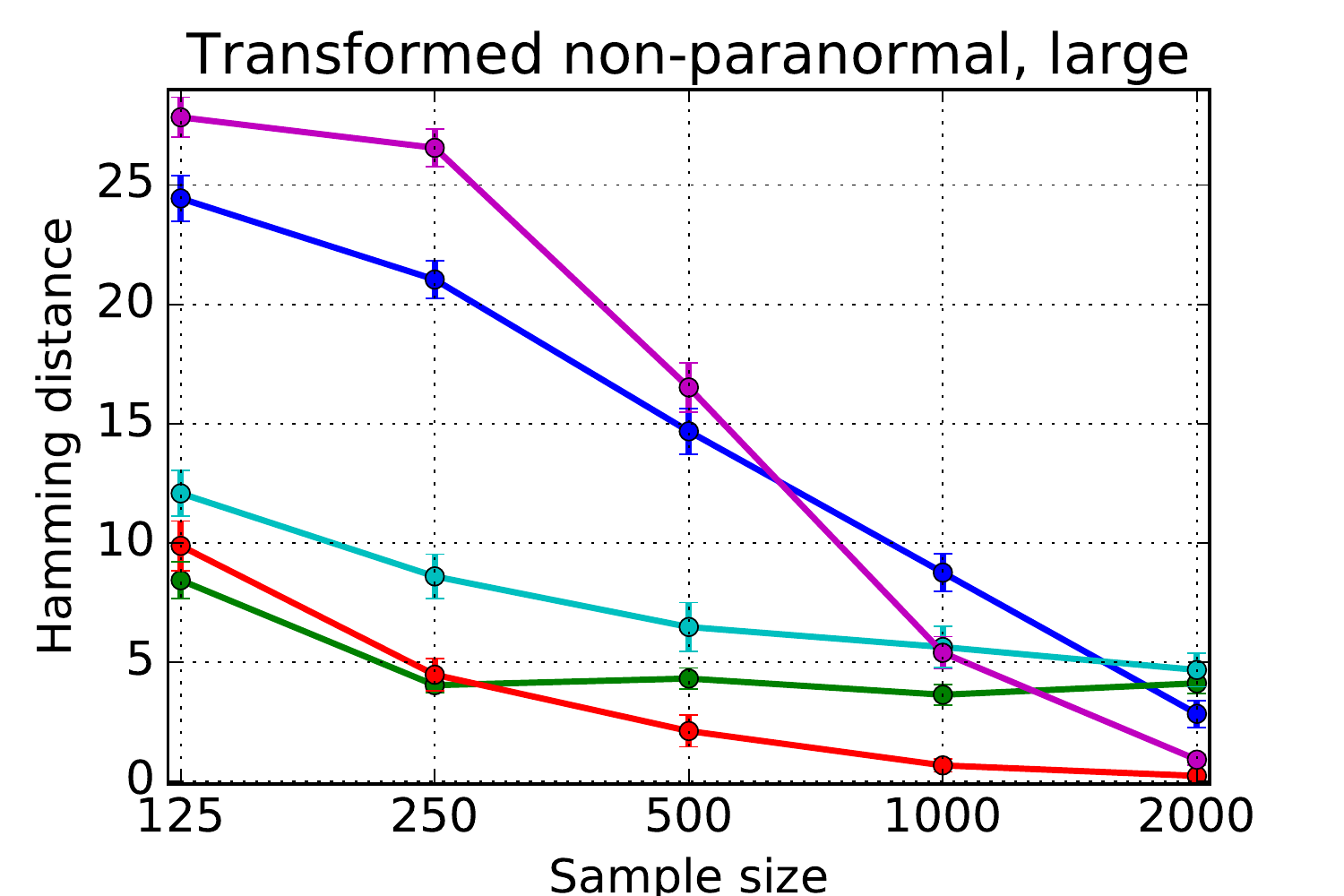}
\includegraphics[width = 0.65\columnwidth]{legend.png}
\end{center}
\caption{Average Hamming distances for random networks}\label{fig:randomNonpara}
\end{figure}

Next, we generated multivariate normal and non-paranormal data from randomly generated graph structures. The graphs were first created by randomly adding an edge between variables with a probability of $3/p$, where $p$ is the number of variables. This implies that the expected number of edges is $3(p-1)/2$. The multivariate normal data was sampled using an R-package 'huge', and the non-paranormal data was created from this by applying a power transformation $X_i \mapsto X_i^3$ to each variable.  The sample sizes of created data sets ranged from $125$ to $2000$. The results are shown in Figure \ref{fig:randomNonpara}. We consider dimension $p = 10$ (the upper row of Figure \ref{fig:randomNonpara}) and $p = 20$ (the lower row). The plots on the left column present the results for the Gaussian data, center column for the non-paranormal and the the right column shows the non-paranormal results when the ECDF transformation is used also with \texttt{knnMI\_AND} and \texttt{fisherZ\_AND}. The results are averages computed from $25$ different graphs.   

Looking at the results, we can see that \texttt{mb} and \texttt{glasso} perform the best when $d = 20$ but worse than others in the smaller dimensional case. \texttt{knnMI\_AND} and \texttt{mb\_auto} perform quite similarly in this experiment, with Hamming distance tending steadily to zero as $n$ increases. If the non-paranormal transformation is applied with \texttt{fisherZ\_AND}, we can see a drastic increase in its performance as evident in the plots on the right column. This suggests that the good performance \texttt{mb}, \texttt{glasso} and \texttt{mb\_auto} can attributed to the used transformation. The performance of our method stays basically the same, which is also in line with the theory as the mutual information is invariant with respect to smooth and invertible transformations on variables, see, for instance, \citep{kraskov2004}. The conclusion from the Gaussian and non-paranormal cases is that many of 
the methods, even the one based on Fisher's $z$-test, seem to behave 
quite well, especially if a reverse transformation is applied. The 
proposed non-parametric method requires more data than some of the other 
methods but seems to eventually converge to the correct network, which 
was to be expected.
\subsection{Large network}
In this setting, we consider a larger network with non-linear dependencies between the variables. The graph is created by combining three seven nodes graphs (depicted in Figure \ref{fig:smallNetwork}) as disconnected components to form a larger $21$ node graph. In each of these independent sub-graphs, data is generated according to non-linear mechanism, as explained in Section \ref{sec:smallNetwork}. The results  averaged from $25$ tests are shown left in Figure \ref{fig:largeNetwork}.
We can see that here our method clearly outperforms the other approaches clearly regardless of the type of noise. The other methods do not seem to be able to capture the structure any better as the sample size is increased. We also considered other noise distributions (Gaussian and uniform) and the results were similar.
\subsection{G-Wishart-Gaussian mixture data}
\begin{figure}[tb]
\begin{center}
\includegraphics[width = 0.31\columnwidth]{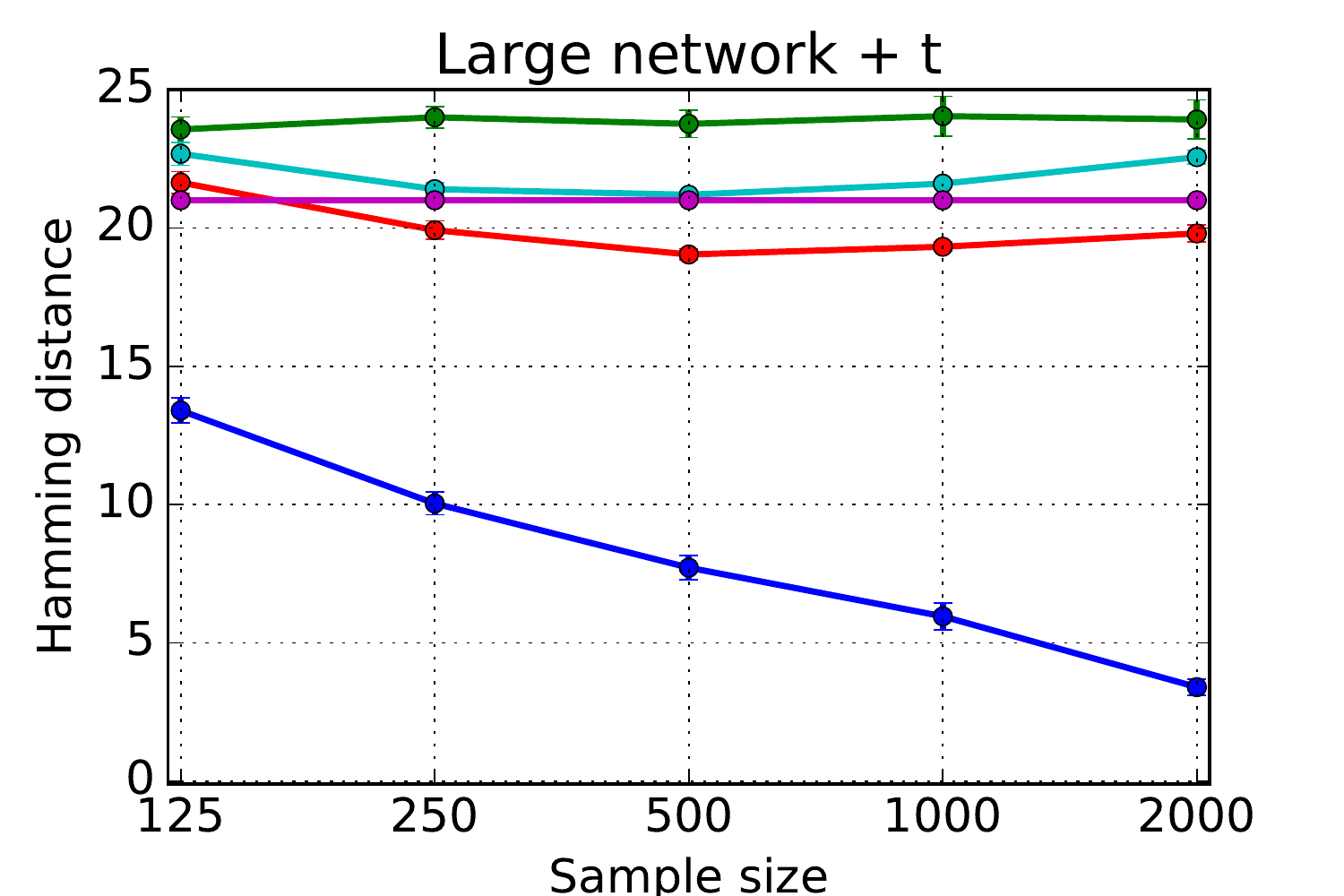}\includegraphics[width = 0.31\columnwidth]{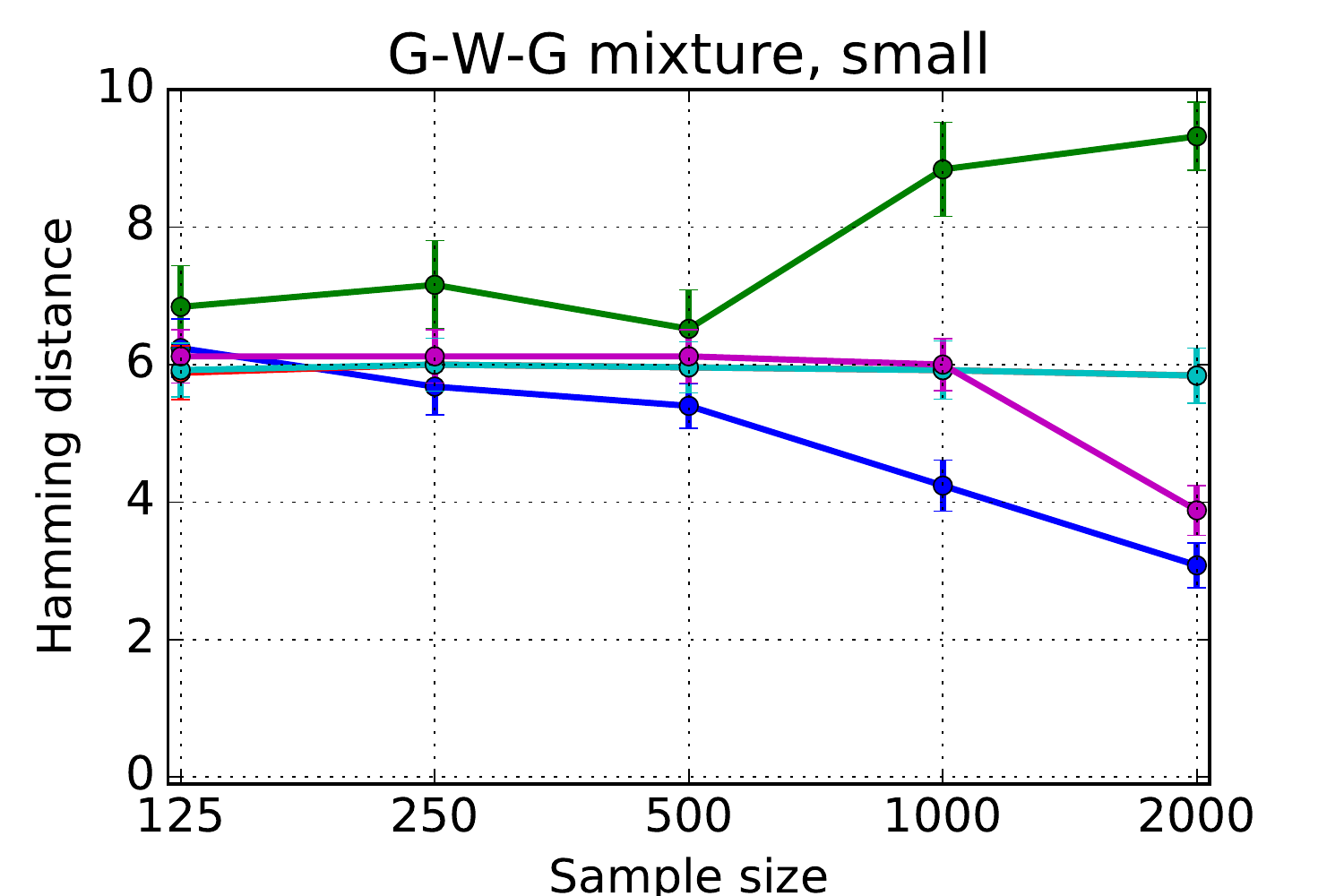}\includegraphics[width = 0.31\columnwidth]{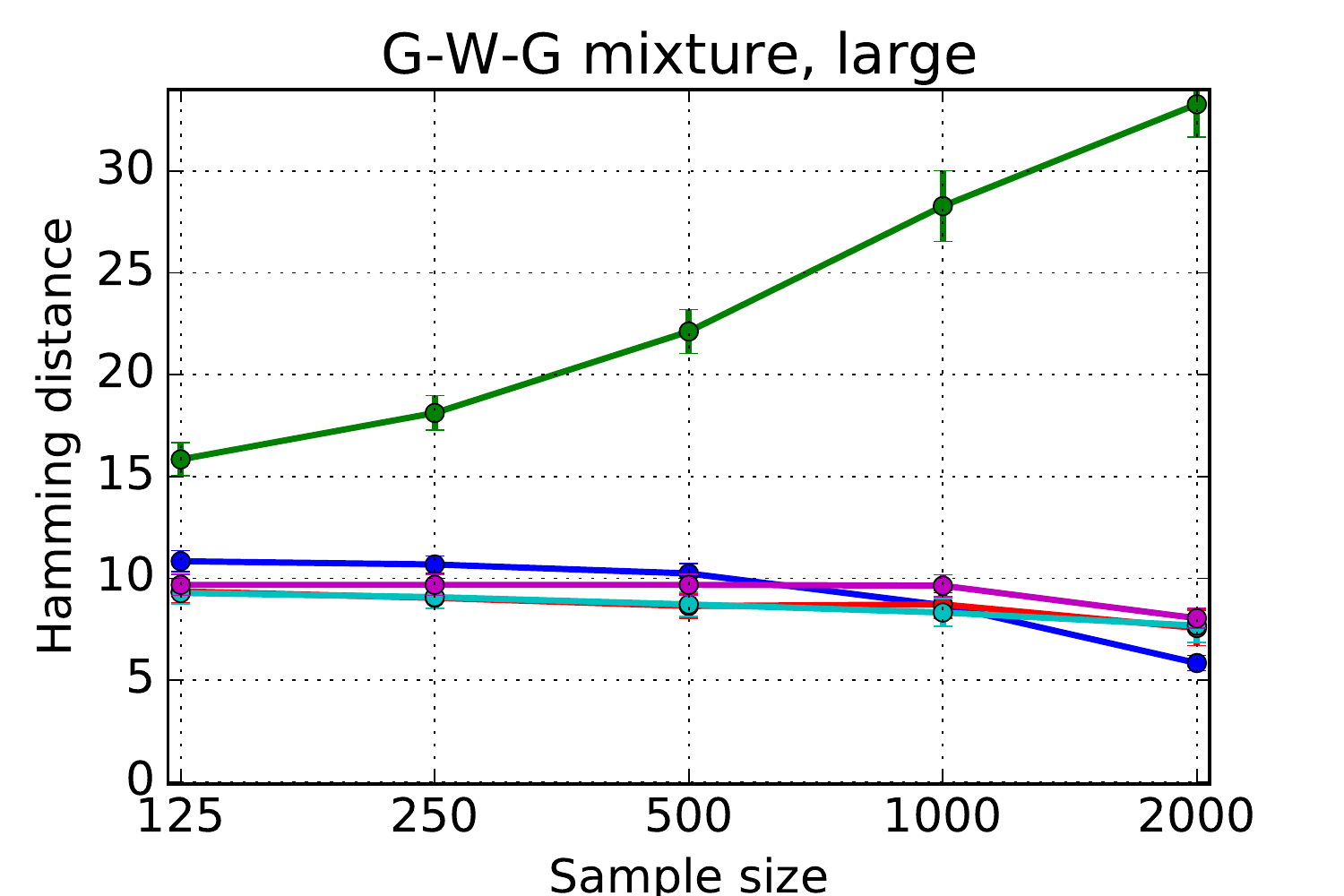}
\includegraphics[width = 0.65\columnwidth]{legend.png}
\caption{Averaged Hamming distances for large network and G-W-G mixture data.}\label{fig:largeNetwork}
\end{center}
\end{figure}

In the final experiment, we start by creating a random undirected graph. Then we sample a matrix from the G-Wishart distribution given this graph. The G-Wishart distribution can be used to sample precision matrices compatible with the conditional independence statements as implied by the graph in the Gaussian graphical model \citep{Lenkoski2011,Lenkoski2013}. Given this precision matrix, we sample one data point from a multivariate normal distribution with zero-mean. This is repeated until we have $n$ samples. The data does not follow a multivariate normal distribution even though each individual point does. We consider dimensions $p = 8$ and $p = 16$ with expected numbers of edges $6$ and $10$, respectively. The degrees of freedom parameter for G-Wishart was set to $3$ and scale matrix had ones on the diagonal, and $0.15$ as every off-diagonal element. R-package 'BDGraph'\footnote{\url{https://CRAN.R-project.org/package=BDgraph}} was used to sample from the G-Wishart distribution. Averaged results from 25 tests are shown in the middle and on the right in Figure \ref{fig:largeNetwork}. 
Here, our method works the best in the small dimensional setting $p = 8$, and similarly to others when $p = 16$. When $n = 2000$ and $p = 16$ our method is the most accurate.
\section{Conclusions}
We have presented an algorithm for distribution free learning of Markov network structures. The algorithm combines previous work on non-parametric estimation of mutual information to an efficient structure learning algorithm in a novel way. 
The \texttt{knnMI\_AND} algorithm consistently outperforms other tested algorithms in structure learning in the case of strongly non-linear dependencies and its performance is robust to non-Gaussian noise.

Even though the Markov blanket searches and permutation tests can be computed in parallel, the computational cost of \texttt{knnMI\_AND} algorithm is noticeably greater than that of other tested algorithms. The nearest-neighbour search is a costly operation, which, especially in the high dimensional case, uses the largest proportion of computation time, even while using efficient metric tree structures. A clear direction for future research is to study if approximate nearest-neighbour searches could by utilized to improve the efficiency while still maintaining the consistent estimation of mutual information.



\newpage
\bibliographystyle{abbrvnat}
\begin{small}
\bibliography{knnmi-NIPS}
\end{small}

\end{document}